\documentclass[letterpaper, 10 pt, conference]{ieeeconf}  

\usepackage{amsmath,amsfonts,amssymb}
\usepackage{algorithm}
\usepackage{float}
\newfloat{algorithm}{t}{lop}
\usepackage{xcolor}
\usepackage{algorithmic}
\usepackage{array}
\usepackage{textcomp}
\usepackage{stfloats}
\usepackage{url}
\usepackage{verbatim}
\usepackage{cite}
\usepackage{booktabs}
\usepackage{multirow}
\usepackage{graphicx}
\usepackage{subcaption}
\usepackage{mathtools,leftindex,tensor,mhchem}

\definecolor{highlander_blue}{RGB}{0,61,165}

\makeatletter
\newcommand\fs@spaceruled{\def\@fs@cfont{\bfseries}\let\@fs@capt\floatc@ruled
  \def\@fs@pre{\vspace{0.5\baselineskip}\hrule height.8pt depth0pt \kern2pt}%
  \def\@fs@post{\kern2pt\hrule\vspace{-0.95\baselineskip}}
  \def\@fs@mid{\kern2pt\hrule\kern2pt}%
  \let\@fs@iftopcapt\iftrue}
\makeatother

\IEEEoverridecommandlockouts                              

\title{\LARGE \bf
An Annotation-to-Detection Framework for Autonomous and Robust Vine Trunk Localization in the Field by Mobile Agricultural Robots
}

\author{Dimitrios Chatziparaschis,$^{1}$ Elia Scudiero,$^{2}$ Brent Sams,$^{3}$ and Konstantinos Karydis$^{1}$
\thanks{$^{1}$~Dept. of Electrical and Computer Engineering,~$^{2}$~Dept. of Environmental Sciences, Univ. of California, Riverside, 900 University Avenue, Riverside, CA 92521, USA;
{\tt\footnotesize\{dchat013, scudiero, karydis\}@ucr.edu}, 
and $^{3}$~Gallo, 600 Yosemite Blvd, Modesto, CA 95354, USA; \tt\footnotesize brent.sams@ejgallo.com.}
\thanks{We gratefully acknowledge the support of NSF (\#CMMI-2046270, \#CNS-2312395, \#CMMI-2326309), USDA-NIFA \#2024-67022-42532, ONR \#N00014-18-1-2252, and the University of California \#UC-MRPI M21PR3417. Any opinions, findings, and conclusions or recommendations expressed in this material are those of the authors and do not necessarily reflect the views of the funding agencies.}
}

\begin{document}

\maketitle
\thispagestyle{empty}
\pagestyle{empty}

\begin{abstract}

The dynamic and heterogeneous nature of agricultural fields presents significant challenges for object detection and localization, particularly for autonomous mobile robots that are tasked with surveying previously unseen unstructured environments. Concurrently, there is a growing need for real-time detection systems that do not depend on large-scale manually labeled real-world datasets. In this work, we introduce a comprehensive annotation-to-detection framework designed to train a robust multi-modal detector using limited and partially labeled training data. The proposed methodology incorporates cross-modal annotation transfer and an early-stage sensor fusion pipeline, which, in conjunction with a multi-stage detection architecture, effectively trains and enhances the system's multi-modal detection capabilities. The effectiveness of the framework was demonstrated through vine trunk detection in novel vineyard settings that featured diverse lighting conditions and varying crop densities to validate performance. When integrated with a customized multi-modal LiDAR and Odometry Mapping (LOAM) algorithm and a tree association module, the system demonstrated high-performance trunk localization, successfully identifying over 70\% of trees in a single traversal with a mean distance error of less than $0.37$\;m. The results reveal that by leveraging multi-modal, incremental-stage annotation and training, the proposed framework achieves robust detection performance regardless of limited starting annotations, showcasing its potential for real-world and near-ground agricultural applications.

\end{abstract}


\section{Introduction}

Precision agriculture increasingly relies on systems operating near the ground, including autonomous robots, to improve real-time field monitoring and enable optimized yield prediction and more sustainable operations with reduced labor and costs~\cite{sparrow2021robots}. 
A key aspect of autonomous robots performing near-ground proximal sensing tasks is their ability to robustly localize plants based on their distinct characteristics, such as tree trunks, enabling the development of per-plant temporal profiles that are essential for both targeted monitoring and comprehensive field assessments~\cite{MORBIDINI2026soilvolcont, hyyppa2020under, niknejad2023phenotyping,scudiero2022neargroundsoilmoisture}. 
However, reliable on-the-go landmark (object of interest) detection remains challenging owing to dynamic environmental conditions, including wind, lighting variability, and interference from dense vegetation~\cite{wosner2021objdetectioninagbenchmark}. 
Therefore, developing methods to ensure robust plant (such as trees and grapevines) detection under changing field conditions is essential for fully realizing the benefits of robot-assisted precision agriculture.

\begin{figure}[!t]
\vspace{6pt}
    \centering
    \includegraphics[width=1\linewidth]{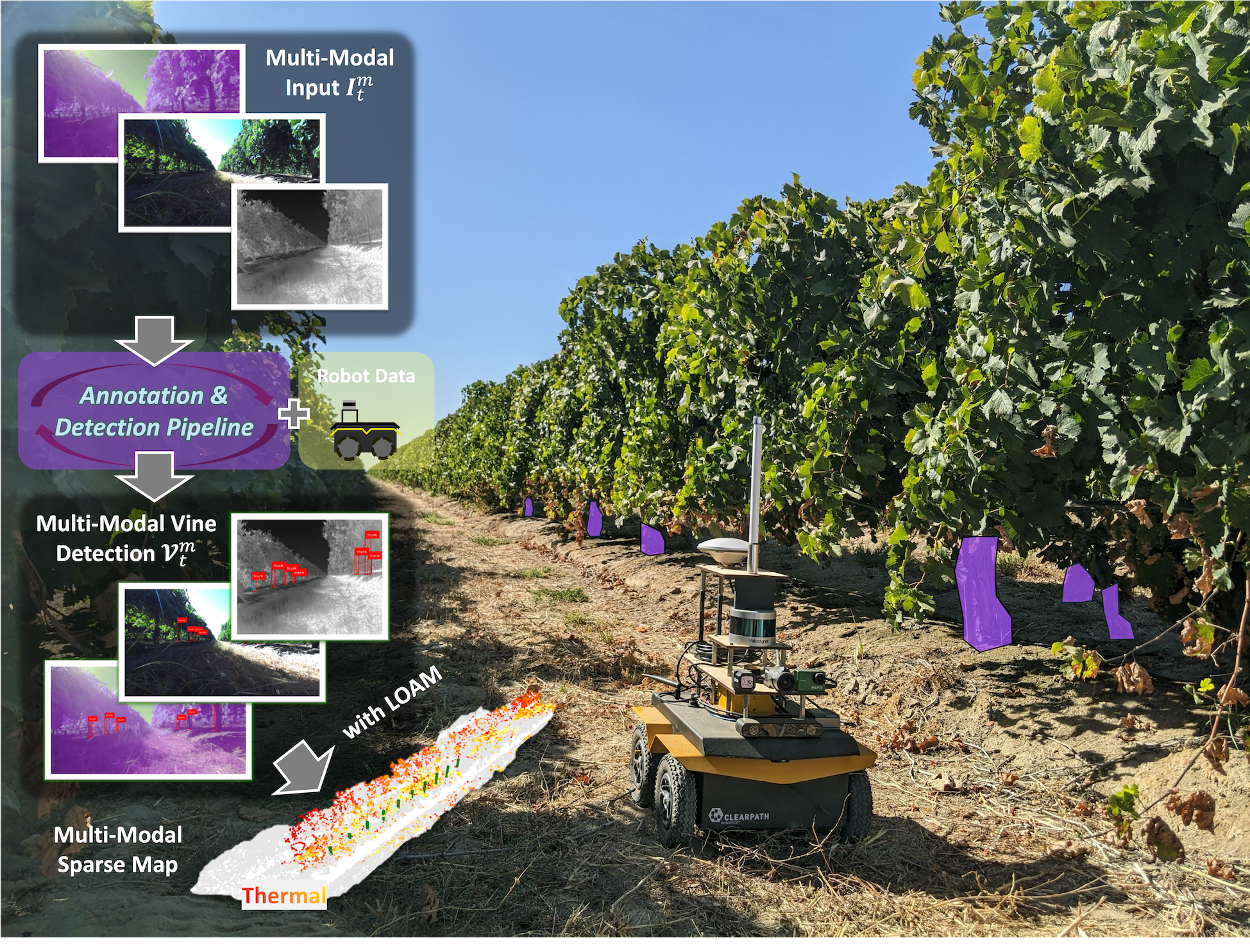}
    \caption{Our annotation-to-detection framework deployed onboard for multi-modal vine trunk detection. Integration with a customized LOAM framework enables concurrent vine trunk localization and the generation of attribute-rich sparse maps.}
    \label{fig:mobile_robot_in_vine_field}
    \vspace{-6pt}
\end{figure}

Multi-modal sensing has proven beneficial for object detection tasks in agricultural and in-the-wild environments, equipping robots with cross-modal capabilities~\cite{parr2021multimodaltoimproveag}. %
The fusion of diverse visual cues that exceed the visible spectrum, along with proximal spatial sensing (such as LiDAR data), can deepen the awareness of robotic operation through attribute-rich mapping, which has been presented in both early and late fusion frameworks~\cite{zamani2023earlylatefusionindetection,gadzicki2020earlyvslatefusion}. 
Recent works, such as~\cite{lu2024multimodaltransformeragdisease}, 
have incorporated textual and context-aware modalities to further enhance model abilities in object detection and classification. 
Reliable frameworks for detection and semantic segmentation have used visual modalities with YOLO~\cite{tian2025yolov12, thumig2024yolov10} and Segment Anything Model (SAM)~\cite{ravi2024sam2,kirillov2023sam1} frameworks, and have been applied in the agricultural domain~\cite{el2025cnninmultimodalag, BADGUJAR2024yoloinagsurvey}.
However, such systems require curated datasets of the given setting for fine-tuning, which are mostly obtained by manual data collection and high-volume human annotation. 
While these datasets ensure robustness in downstream tasks, their creation remains a bottleneck for field deployment and may limit generalizability across different environments~\cite{wosner2021objdetectioninagbenchmark, el2025cnninmultimodalag}.

Semi-Supervised Learning (SSL) has been introduced and applied in cases where large-scale unlabeled data can be easily obtained,
via pseudo-labeled data integration~\cite {huang2022fewshotandselfsupervisedsurvey}. 
Such approaches have proven effective in the agricultural domain for object detection tasks~\cite{tseng2023semisupervised},
whereas they leverage multi-modal information to yield performance gains in detection accuracy~\cite{deva2019pseudothermalsemi}.
Conversely, Few-Shot Learning (FSL)~\cite{xin2024fewshotlearning,wang2020fewshotlearning} has demonstrated remarkable results when smaller datasets are available, mainly in classification tasks~\cite{ragu2022fewshotlearninginag},
incorporating unseen support datasets by employing similarity criteria during training. Although detection systems have been widely demonstrated in the field, cross-modal approaches must be further examined, particularly when operating in real-time and in-the-wild environments. 

In this work, we present a multi-modal annotation-to-detection framework to develop a robust model for cross-modal object detection in the field when limited or no labeled data are available. 
Our framework integrates a coupled twofold pipeline: 1) utilizing early-stage pseudo-labels from a frozen semantic annotator~\cite{kirillov2023sam1} and association through spatial and visual modalities fusion, and 2) a multi-stage training procedure that uses prior detection knowledge~\cite{thumig2024yolov10} to enrich training datasets and enhance multi-modal detection on-the-go with minimal human intervention.
Our cross-modal annotation-to-detection system was deployed on a mobile agricultural robot and evaluated in unseen and vegetation-dense vineyards, specifically for vine trunk detection (Fig.~\ref{fig:mobile_robot_in_vine_field}). 
Extensive field evaluation showcased our detector's performance in vine trunk detection, while demonstrating robust tree localization properties when combined with a multi-modal modified version of the AG-LOAM algorithm~\cite{teng2025adaptiveloam} and an underlying tree trunk association framework~\cite{chatziparaschis2024onthego}.
The main contributions of this work are as follows: 

\begin{itemize}
    \item An early-fusion, cross-modal annotation pipeline that generates object-of-interest masks enriched with spatial information and multi-modal attributes (e.g., thermal).
    \item A stage-incremental detection pipeline utilizing prior knowledge to iteratively refine and include precise pseudo-labels to achieve robust detection performance while minimizing laborious human intervention.
    \item Accurate in-field vine localization, integrating a multi-modal enhanced LOAM framework~\cite{teng2025adaptiveloam} with a tree-association algorithm~\cite{chatziparaschis2024onthego}, to enable generation of feature-rich sparse point maps accompanied by precise tree detections.
\end{itemize}

\section{Related Works}

Multi-modal and data-driven predictive models have shown a positive impact on agricultural applications, spanning from yield prediction to crop detection tasks~\cite{el2025cnninmultimodalag}.
Dataset creation and correct annotation are essential for performing any type of data fusion~\cite{mitchell2010imagefusion}, particularly in tasks where more than one sensing modality is employed~\cite{feng2021multimodalobjdetsurvey}. Open datasets such as Treescope~\cite{cheng2024treescope} have been released to advance autonomous mapping and investigate tree phenomics~\cite{houle2010phenomics}, with a particular focus on aerial robotic platforms.  

In vineyard settings, the VineSet dataset was published~\cite{aguiar2020trunkdetectionmdpi} to support multi-modal vine trunk detection, as well as to facilitate grape bunch detection across various growth stages~\cite{aguiar2021grapebunchdetection}. An initial evaluation was conducted using a mobile robotic platform for vine data acquisition and manual annotation, followed by the deployment of a Tensor Processor Unit (TPU) module for real-time trunk detection~\cite{aguiar2020trunkdetection}. 
In another study, Magalh{\~a}es~$et~al.$~\cite{magalhaes2023benchmarkingedgefordetection} employed VineSet to evaluate the real-time performance of various heterogeneous detection platforms. However, while public datasets are vital for model training, operating environments and conditions may differ among robotic applications, leading to poor performance and cross-domain adaptation~\cite{oza2024unsuperviseddomainadapt}. 
Therefore, a more direct, in situ approach is required to leverage the existing knowledge of capable detectors, incorporating multi-modal reasoning while minimizing human intervention.

Vine trunk detection has been a primary and integral component of robotic research, extending it to localization and mapping. 
Specifically, Papadimitriou~$et~al.$~\cite{papadimitriou2022loopclosureinvines} utilized field uniformity to obtain and spatially characterize valuable visual trunk features to perform loop-closure detection. Slavi{\v{c}}ek P.~$et~al.$~\cite{slavivcek2024annotationvines} presented a YOLOv5 framework that was trained and used to obtain enhanced trunk annotations on available open datasets (including VineSet~\cite{aguiar2021grapebunchdetection}). Specifically, by using a new dataset, a student-teacher network approach for vine trunk detection was employed, and post-filtering with human intervention was applied to exclude faulty detections. Orchard trunk and shrub detection for robotic obstacle avoidance was also demonstrated in~\cite{xiangyang2024trunkshrubdatasetanddetect}, with manual collection and labeling of images. Liu Y.~$et~al.$~\cite{liu2023trunkdetectionyolov7} presented a YOLOv7-based trunk detection for $Camellia$ $oleifera$ fruit trees, improving the backbone feature extraction method and training loss for the specific task. Fruit counting on vertical fruiting-wall trees with trunk detection and tracking was also demonstrated by Gao F.~$et~al.$~\cite{gao2022fruitcountingapples} having initially manual annotation and counting apple fruits and trunks. In all these cases, manual annotations were made for specific tasks, often requiring considerable effort from annotators. Cao Z.~$et~al.$~\cite{cao2024orchardlineextryolov8} collected vine trunk datasets and used a Mix-Shelter method for data augmentation with a customized YOLOv8
detector to extract tree lines for safe navigation in orchards. In our work, we present an annotation pipeline as an integral component of our training procedure that aids diverse data augmentation to obtain an accurate in-domain multi-modal detector for vine trunk detection.

\begin{figure*}[!t]
\vspace{6pt}
    \centering
    \includegraphics[width=1\linewidth]{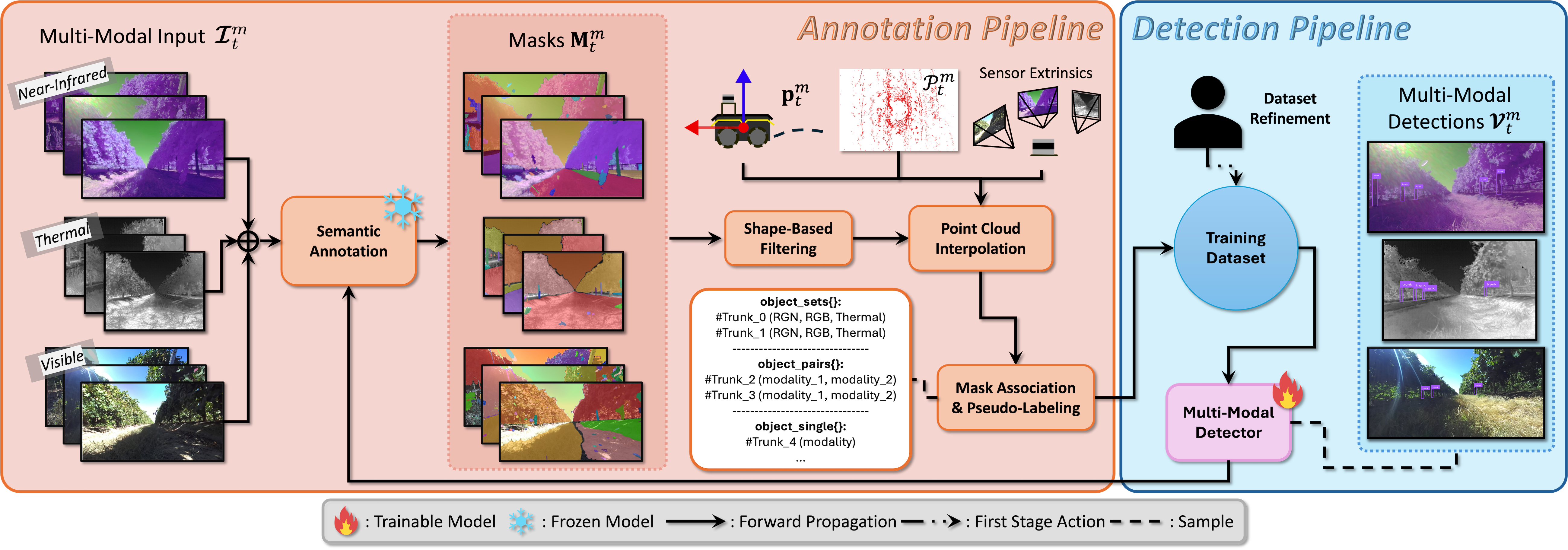}
    \caption{\textbf{Our proposed annotation-to-detection framework for multi-modal vine trunk detection.} The annotation pipeline initiates a tree trunk dataset in the form of annotation masks $\textbf{M}_t^m$ by combining multi-modal input ($\pmb{\mathcal{I}}^v_t$,~$\pmb{\mathcal{I}}^n_t$,~$\pmb{\mathcal{I}}^{\tau}_t$) and available spatial information during the robot's operation. 
    The detection pipeline is coupled with the annotated (pseudo-labeled) dataset to iteratively refine and augment the available data, towards training a robust multi-modal vine $\pmb{\mathcal{V}}_t^m$ detector.}
    \label{fig:system_overview}
    \vspace{-7pt}
\end{figure*}

\section{System Description}

\subsection{Sensing Modalities and Specifications}

We deployed a Clearpath Jackal mobile robot (Fig.~\ref{fig:mobile_robot_in_vine_field}) equipped with three fixed-in-place imaging sensors for detection, spanning from visible, Near-Infrared (NIR), to Long-Wave Infrared (LWIR) spectra. 
The visible modality (RGB) was obtained using a Stereo Labs Zed2i depth camera utilizing the left camera's $720p$ resolution imagery footage. 
A Mapir Survey3 digital camera was used to capture the NIR data (i.e. through Red-Green-NIR) at a resolution of $720p$, and a FLIR ADK sensor was used to acquire thermal data at $640\times512$ pixel resolution. 
Along with the imaging sensors, a Velodyne VLP-16 LiDAR was employed for sparse 3D point cloud generation, which was later used for both mapping and cross-camera detection correspondences. 
The imaging sensors were set to a $10$Hz publishing rate to match the 3D LiDAR operating rate. 

The state estimation of the robot in the field was derived from wheel odometry data to determine its pose within a local area. 
In addition, RTK-GNSS data were used to acquire fixed $cm$-level global positions using a nearby established RTK-GNSS station. 
By incorporating the robot's heading, we can apply a forward geodesic transformation to reference each vine detection to the global frame with WGS84 longitude and latitude coordinates. 
In this way, we generated globally georeferenced AG-LOAM maps from the field, enriched with multi-modal information and the final tree trunk detections.

\subsection{Data Synchronization and Point Cloud Integration}

It is important to ensure accurate sensor synchronization and data interpolation to enable reliable detection, particularly when multiple modalities are used. 
Let $\pmb{\mathcal{I}}^m_t \in \mathbb{R}^{H\times W\times C}$ be the captured image from the $m$ sensing modality with $m\in\{visual, nir, thermal\}$.
Each modality might have a different number of $C$ channels, such as the single in thermal image $\pmb{\mathcal{I}}^{\tau}_t$. Initially, a set of images $\pmb{\mathcal{I}}^m_t$ at time $t$ is obtained using soft synchronization with respect to the common clock of the robot's ROS environment. 
Together with the set of images ($\pmb{\mathcal{I}}^v_t$,~$\pmb{\mathcal{I}}^n_t$,~$\pmb{\mathcal{I}}^{\tau}_t$), the corresponding robot pose $\mathbf{p}^m_t$ is stored as a transformation matrix in $SE(\mathit{3})$ with respect to the robot's starting position at time $t$. 

Let $\mathcal{P}_{t'}\in \mathbb{R}^{3}$ be the captured point cloud at time $t'\ge t\in\mathbb{R}^{+}$. 
To obtain an accurate point cloud projection for each image $\pmb{\mathcal{I}}^m_t$, we interpolate $\mathcal{P}_{t'}$ to the previous time step $t$ by using the relative transformation from $\mathbf{p}_{t'}$ to $\mathbf{p}^m_t$, owing to robotic movement. Thus, for or each $\pmb{\mathcal{I}}^m_t$ we have,
\begin{gather*}
    \mathcal{P}^m_{t} = (\mathbf{p}^m_t )^T\cdot \mathbf{p}_{t'} \cdot \mathcal{P}_{t'}
    = \leftindex^{t'}_{t\hspace{0.05cm}}{T} \cdot \mathcal{P}_{t'}  
\end{gather*}
be the interpolated $\mathcal{P}_{t'}$ from $t'$ to $t$ timestamp, and $\leftindex^{t'}_{t\hspace{0.05cm}}{T}$ is the relative transformation matrix. 
Since all sensors are affixed to the robot's chassis, their inner transformations are static. 
We used the KALIBR calibration tool to calculate each sensor's camera matrix $K_m$ and the ACFR library~\cite{tsai2021optimising} 
for their extrinsic calibration relative to the VLP-16 LiDAR.
Thus, having the transformed point clouds $\mathcal{P}^m_{t}$, we can use the sensor intrinsics and relative poses to project the points on their image plane of $\pmb{\mathcal{I}}^v_t$, $\pmb{\mathcal{I}}^n_t$, and $\pmb{\mathcal{I}}^{\tau}_t$, separately.

\section{Proposed Annotation-to-Detection Framework}

\subsection{Annotation Pipeline and Pseudo-Labeling}

\begin{figure}[!t]
\vspace{6pt}
    \centering
    \includegraphics[width=\linewidth]{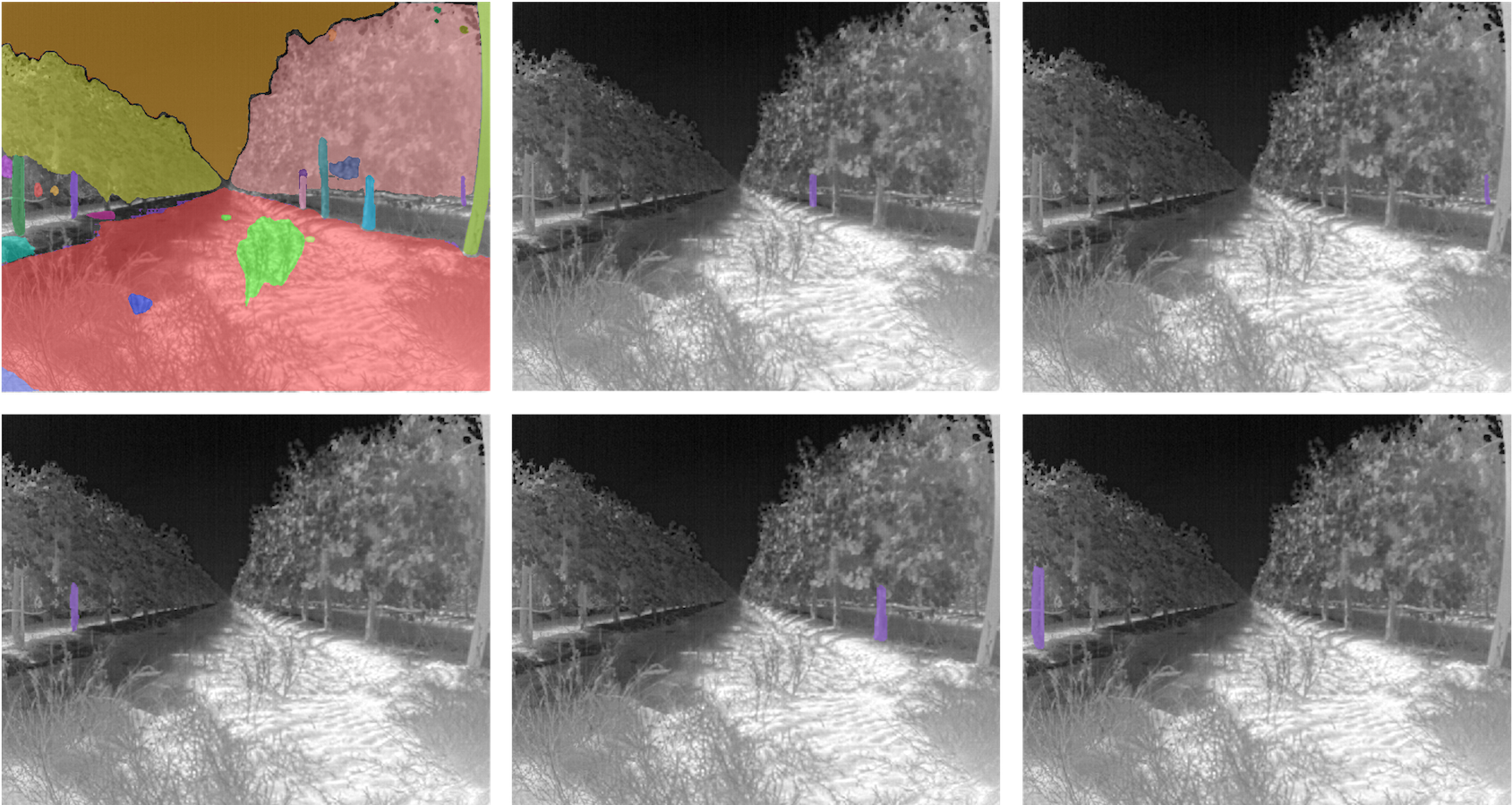}
    \caption{Generated semantic masks $\mathbf{M}_t^{\tau}$, based on a thermal image $\pmb{\mathcal{I}}^{\tau}_t$ input, by using SAM~\cite{kirillov2023sam1} and our shape-based mask filter.}
    \label{fig:SAM_filtered_detections}
\end{figure}

The primary purpose of the annotation pipeline (Fig.~\ref{fig:system_overview}) is to acquire partial but valuable labels of the object of interest, herein vine trunks, to create a multi-modal training set. 
We initially used the SAM annotator~\cite{kirillov2023sam1} on the accumulated images from all modalities $m$ to obtain semantic masks $\mathbf{M}^m_t = \{\mathcal{M}^m_0,\mathcal{M}^m_1,...,\mathcal{M}^m_i\}$ with $\mathcal{M}^m\in\{0,1\}^{H\times W}$.
To acquire vine trunk masks over the $\mathbf{M}^m_t$ set, we applied a rectangular shape-based mask filter by extracting the contours of the semantic masks and evaluating whether the number of vertices was greater than two (holding for quadrilateral shapes). 
By checking the height-to-width contour ratio, we filtered standing rectangle-based masks and considered potential vine trunk masks in the field of view. 
Figure~\ref{fig:SAM_filtered_detections} shows an example of SAM detections on a thermal $\pmb{\mathcal{I}}^{\tau}_t$ capture. 
Overlapping masks with an Intersection over Union (IoU) greater than 0.5, or if they occupy more than 40\% or less than 5\% of the image frame size, are discarded from detection.

\begin{figure}[!t]
\vspace{6pt}
    \centering
    \includegraphics[width=\linewidth]{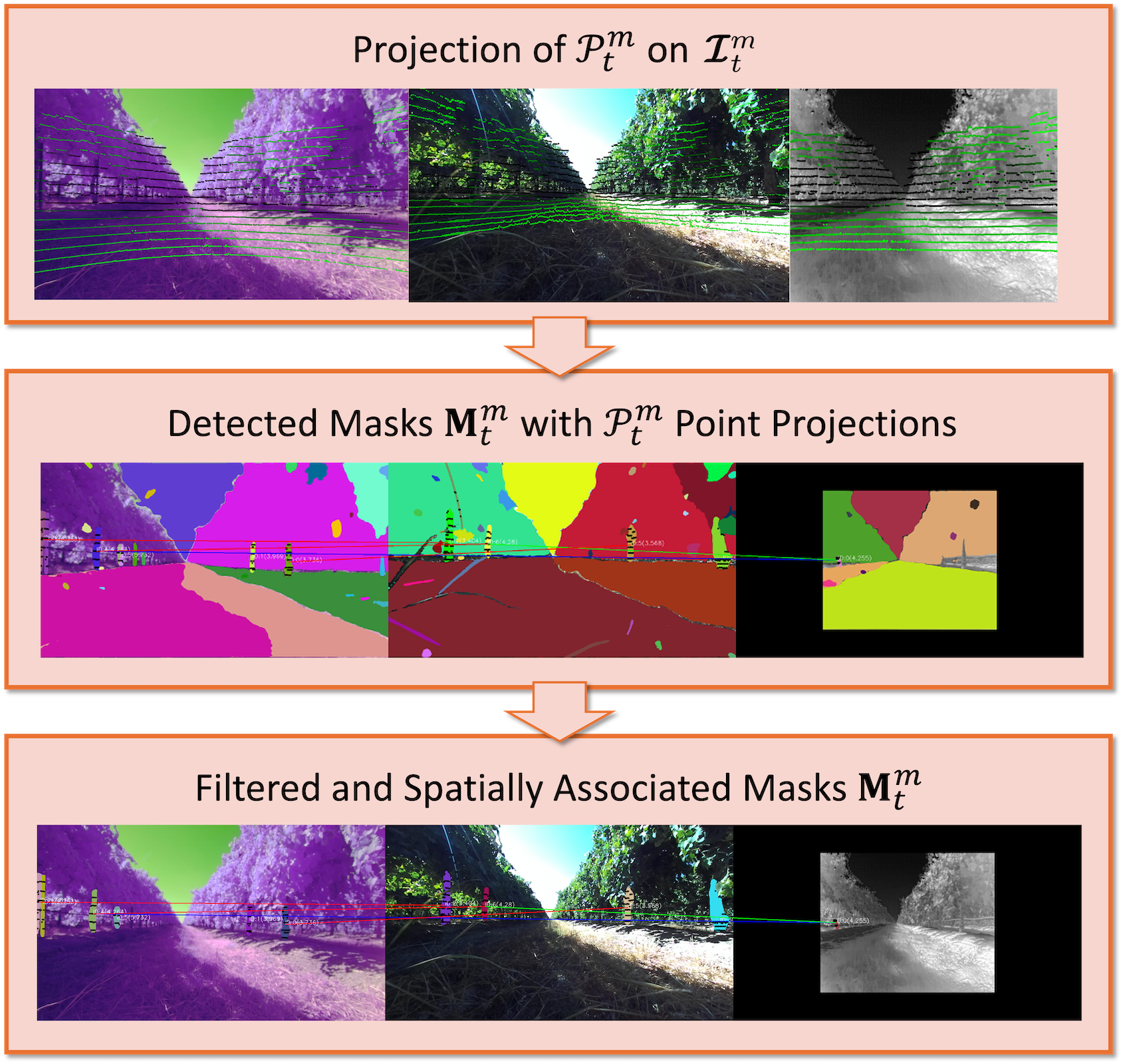}
    \caption{\textbf{Mask association and pseudo-labeling}. By employing the interpolated point clouds $\mathcal{P}_t^m$ and filtered masks $\mathbf{M}_t^m$, the vines are spatially associated to determine if they represent the same object across different modalities.}
    \label{fig:multi_modal_trunk_correlation}
    \vspace{-6pt}
\end{figure}

As the filtered 2D candidates were acquired from each modality, we proceeded with their spatial association using the corresponding interpolated $\mathcal{P}^m_{t}$ point clouds. By projecting $\mathcal{P}^m_{t}$ onto images $\pmb{\mathcal{I}}^{m}_t$, we form sparse 3D positions for each mask in $\mathbf{M}^m_t$.
The centroid points of all lifted masks $\mathbf{M}^m_t$ are computed after we filter out point outliers by statistically removing the ones with distance deviation greater than $2\sigma$ from the distance mean (i.e. in cases of detection of edge pixels on the mask and/or from LiDAR points that were captured behind the observed object).
By iterating over all modalities, we determine the closest cross-modal mask pairs or sets given their 3D centroid positions by having less than $10~cm$ in 3D distance. We form sets of $\{\mathbf{M}^u_t,~\mathbf{M}^n_t,~\mathbf{M}^{\tau}_t\}$ that correspond to the same detected object across all or partial modalities $\{\pmb{\mathcal{I}}^u_t,~\pmb{\mathcal{I}}^n_t,~\pmb{\mathcal{I}}^{\tau}_t\}$ at time $t$. Figure~\ref{fig:multi_modal_trunk_correlation} shows the intermediate steps for 3D vine association and labeling in our annotation pipeline.

\subsection{Detection Pipeline and Multi-Stage Training}

\begin{figure*}[!t]
\vspace{6pt}
     \centering
     \begin{subfigure}{0.24\textwidth}
         \includegraphics[width=\linewidth]{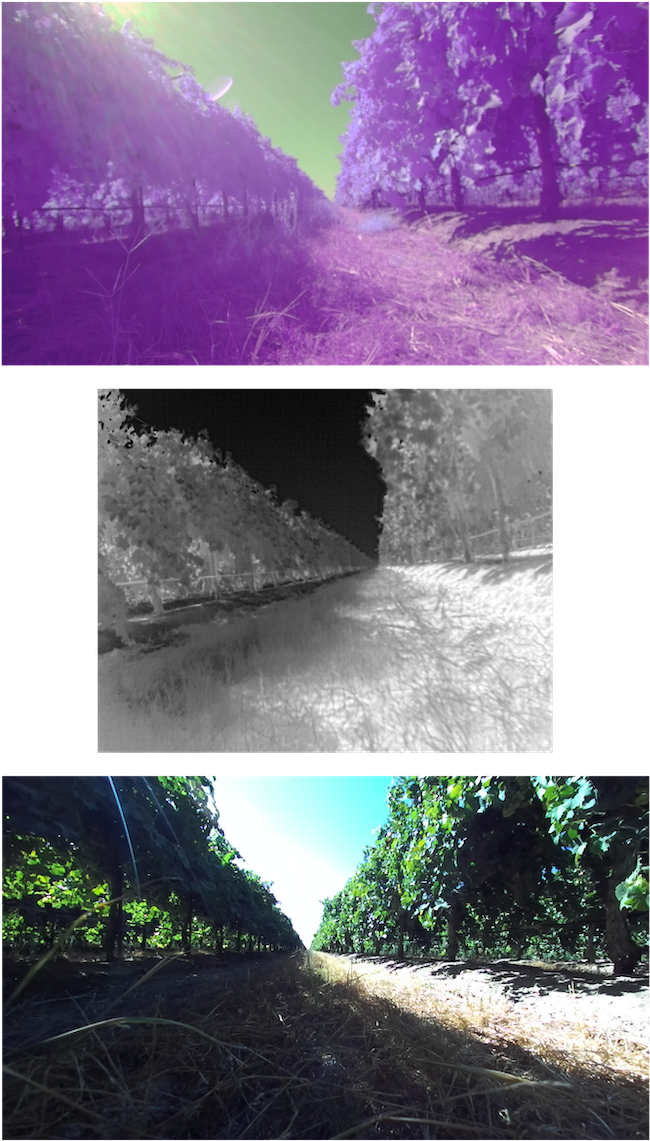}
         \caption{Raw $\pmb{\mathcal{I}}^{m}_t$ Inputs}
     \end{subfigure}
     \begin{subfigure}{0.24\textwidth}
         \includegraphics[width=\linewidth]{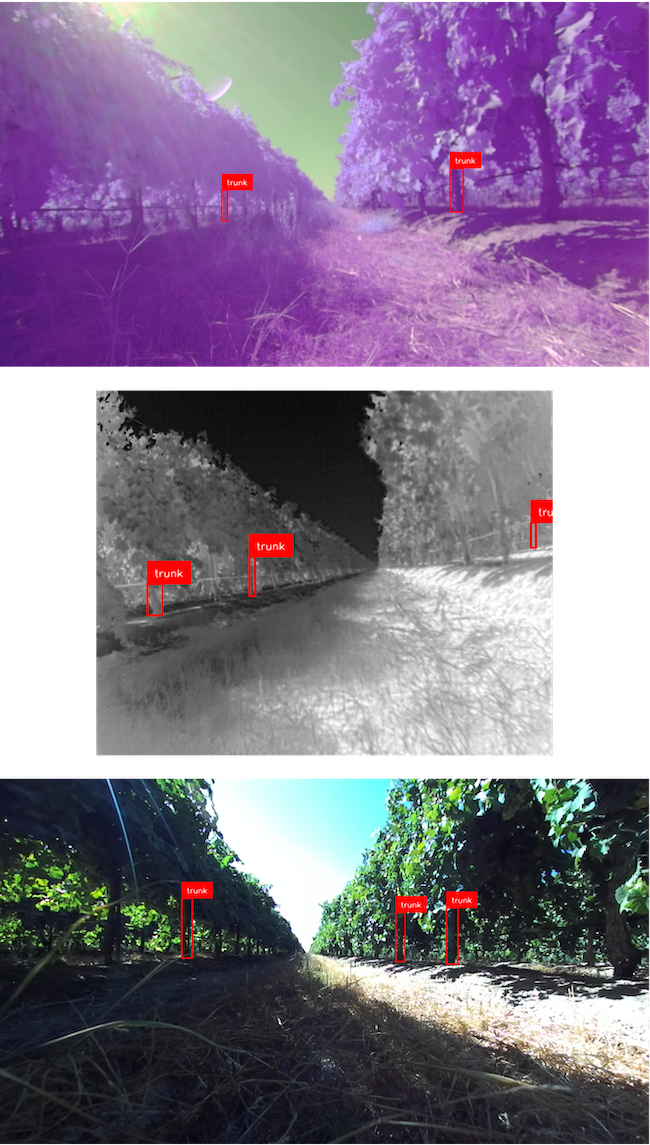}
         \caption{$\pmb{\mathcal{V}}^m_t$ with $\mathcal{D}_{S}$}
         \label{fig:dsam}
     \end{subfigure}
     \begin{subfigure}{0.24\textwidth}
         \includegraphics[width=\linewidth]{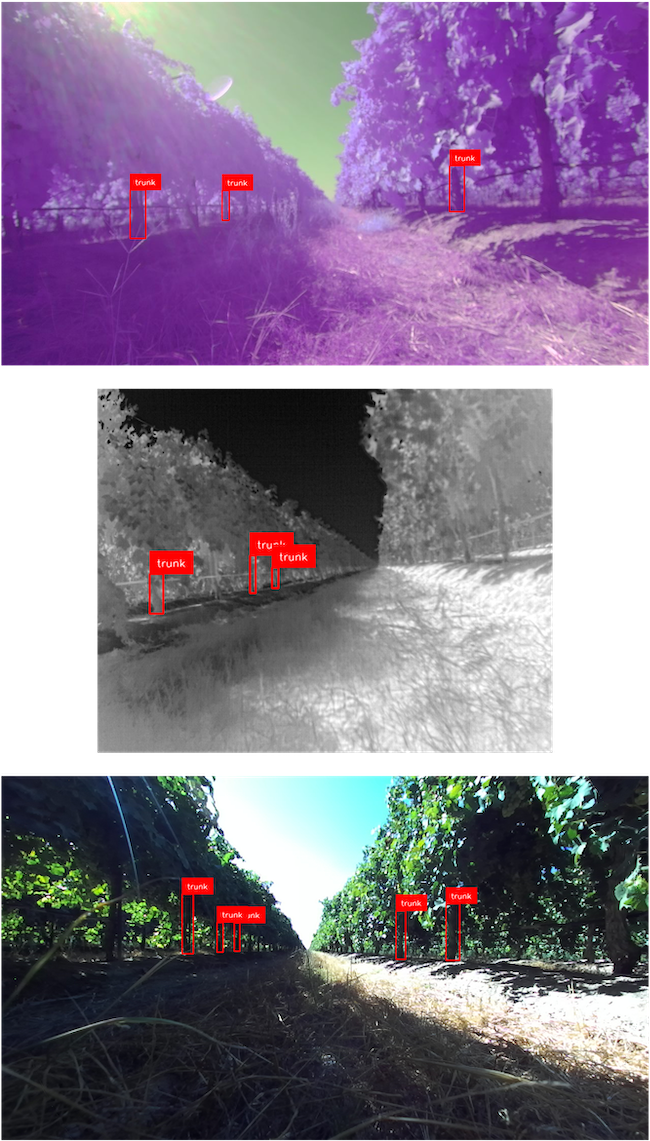}
         \caption{$\pmb{\mathcal{V}}^m_t$ with $\mathcal{D}_{S^+}$}
         \label{fig:dpsd}
    \end{subfigure}
    \begin{subfigure}{0.24\textwidth}
         \includegraphics[width=\linewidth]{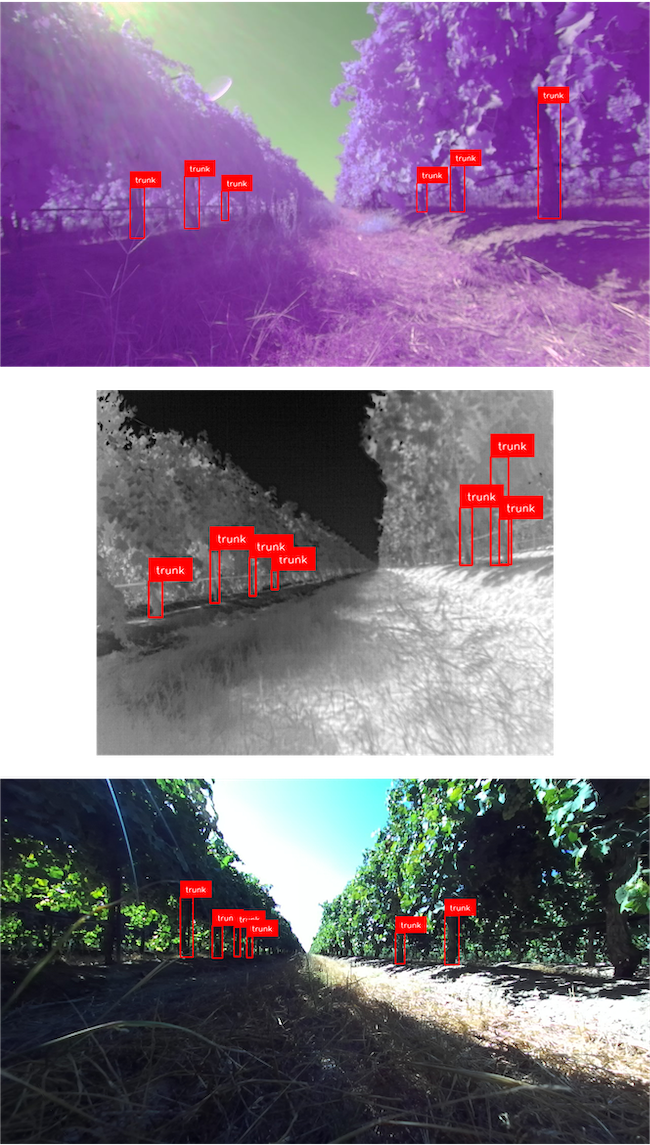}
         \caption{$\pmb{\mathcal{V}}^m_t$ with $\mathcal{D}_{T}$}
         \label{fig:dext}
     \end{subfigure}
     \caption{\textbf{Evolution of vine trunk detection.} By transitioning from the $\mathcal{D}_{S}$ model, trained on the partially annotated dataset (b), to $\mathcal{D}_{T}$ model, which incorporates the enriched pseudo-labeled dataset (d), the detector achieves superior performance across all input modalities.}
\label{fig:detectors_comparison_unseen}
\end{figure*}

The key goal of the detection pipeline (Fig.~\ref{fig:system_overview}) is to gradually train a multi-modal detector $\mathcal{D}$ to obtain vine trunks $\pmb{\mathcal{V}}^m_t$ from any given modality, namely $\mathcal{D}(\pmb{\mathcal{I}}^m_t)\rightarrow\pmb{\mathcal{V}}^m_t$ as instance segmentations. 
Thus, by prioritizing high precision and minimizing false positives, our system iteratively integrates pseudo-labels of detected vines$-$starting with partially annotated data$-$during the multi-stage training procedure to build an accurate detection performance. 
For data collection, we deployed our system in Cabernet Sauvignon vineyards at Gallo Winery ($36^\circ 49'49.4''N$, $120^\circ12'36.7''W$).

Initially, our annotation pipeline generated a small tree trunk dataset with 100 image sets of partial vine trunk correspondences using the SAM annotator. 
The multi-modal dataset was split into training, evaluation, and testing subsets following the 70-20-10 ratio, ensuring an equal ratio among modalities. 
For the detector, we used the pretrained YOLOv10n~\cite{thumig2024yolov10} model and parsed images from all modalities to enable the development of cross-modal properties. 
Only in the first step was a human annotator required to remove false annotations in the training set and complete missing data from the evaluation and testing sets. Notably, the manually curated testing set was used to evaluate our detector's performance among the upcoming training stages, alleviating any further human intervention. 

\begin{table}[!h]
\centering
\caption{Evaluation Stages of Detector $\mathcal{D}$ Aiming Low-False-Positive Vine Trunk Detection}
\vspace{-0pt}
\begin{tabular}{c@{\hspace{0.5em}}c@{\hspace{0.5em}}c@{\hspace{0.5em}}c@{\hspace{0.5em}}c@{\hspace{0.5em}}}
\toprule 
Metrics             & {$\mathcal{D}_{S}$}                                                 & {$\mathcal{D}_{S^+}$}    & {$\mathcal{D}_{T}$}       \\\midrule 
Precision           &  0.02             & 0.16  & \textbf{0.83}           \\
Recall          & 0.14            & 0.35     & \textbf{0.53}       \\
mAP$_{50}$           & 0.17             & 0.37        & \textbf{0.55}     \\
mAP$_{50:95}$           & 0.08              & 0.19  & \textbf{0.26}         \\ \bottomrule
\end{tabular} \label{tab:yolov10_metrics}
\end{table}

Table~\ref{tab:yolov10_metrics} presents the testing results for unseen multi-modal images. 
As the first-stage detector $\mathcal{D}_{S}$ was trained on images with limited vine trunk annotations (i.e. filtered SAM masks), lower precision and recall scores were observed. 
Although the resulting detector, $\mathcal{D}_{S}$, exhibits limited vine detection capabilities, we leverage its high-confidence predictions in the second iteration to include new trunk annotations within the existing training set. 
This iterative refinement yielded $\mathcal{D}_{S^+}$, with a 60\% increase in both recall and precision. 
In the last step, the refined $\mathcal{D}_{S^+}$ model was used to annotate 1500 newly captured multi-modal image sets via the annotation pipeline and repeat training, prioritizing detection precision. 
The new detector, $\mathcal{D}_{T}$, has the best performance, 
scoring $0.83$ in precision and $0.53$ in recall, as it was trained on the extended training set. 
Because the detection pipeline prioritizes the reduction of false positives to ensure the veracity of automated pseudo-labeling, it produces fewer but more accurate detections across modalities, with lower mAP scores as a trade-off. 
Figure~\ref{fig:detectors_comparison_unseen} illustrates the detection performance of  $\mathcal{D}$ models across all training stages on unseen images. 
We maintained a consistent configuration across all training steps, utilizing a batch size of 32 images for 100 epochs and the pretrained YOLOv10n model as the primary detection backbone. Model evaluation and training were supported by a combination of local computational resources and the National Research Platform (NRP) at UC San Diego.

\begin{table*}[!t]
\centering
\caption{Evaluation of Vine Trunk Localization}
\vspace{-0pt}
\begin{tabular}{c@{\hspace{0.5em}}|c@{\hspace{0.5em}}c@{\hspace{0.5em}}c@{\hspace{0.5em}}|c@{\hspace{0.5em}}}
\toprule
Experiment & & Single-Row with $10$ Trees &  & Dual-Row with $7\times7$ Trees     \\\midrule 
Metrics       & $1st-5th$ Tree & $5th-10th$ Tree & Total & Total      \\\midrule 
$L^2$ Distance Error          & 0.32 m & 0.24 m  & 0.28 m & 0.37 m                              \\
MAE$_{<0.5}$          & 0.05 m & 0.11 m  & 0.09 m & 0.06 m                              \\
RMSE$_{<0.5}$         & 0.05 m & 0.12 m  & 0.10 m & 0.07 m                              \\
Recall$_{<0.5}$       & 60\% & 80\%   & 80\%  & 71\%                                \\
Total Tree Detections  & 4 out of 5 & 4 out of 5  & 8 out of 10 & 10 out of 14           \\ 
\bottomrule
\end{tabular} \label{tab:field_evaluation}
\vspace{-0pt}
\end{table*}

\section{Experimental Evaluation}

To evaluate vine trunk detection and localization, we deployed our system across two unseen fields at Gallo Vineyards and conducted surveys between straight-line tree rows. 
By maintaining westward heading during post-noon hours, we examined our system's ability in vine trunk detection and localization across different numbers of trees and sunlight conditions within the rows. 
To derive a sparse 3D map of the surveyed vineyard, we modified the AG-LOAM~\cite{teng2025adaptiveloam} to register information from the additional sensing modalities during LiDAR mapping. 
Along with the mapping system, we integrated a multi-modal tree localization module~\cite{chatziparaschis2024onthego} for on-the-go and by-tree information tracking using our online trunk detections as an input. 
The positions of the detected trees were evaluated against the surveyed ground truth. 
The localization results are presented in Table~\ref{tab:field_evaluation}. 

\subsection{Single-Row Vine Detection}

In this experimental setup, we evaluated the system on a $20\times5~m^2$ vineyard row containing 10 vine trees located to the right of the robot's trajectory. 
As a single tree row, all vines were uniformly exposed to sunlight, yielding a consistent lighting scenario. 
During the survey, the robot maintained a constant linear speed of 0.5$~m/s$ and captured multi-modal data along the path, containing vines at varying distances, angles, and partially occluded instances.

\begin{figure*}
\centering
\begin{subfigure}{.4\textwidth}
  \centering
  \includegraphics[width=1\linewidth]{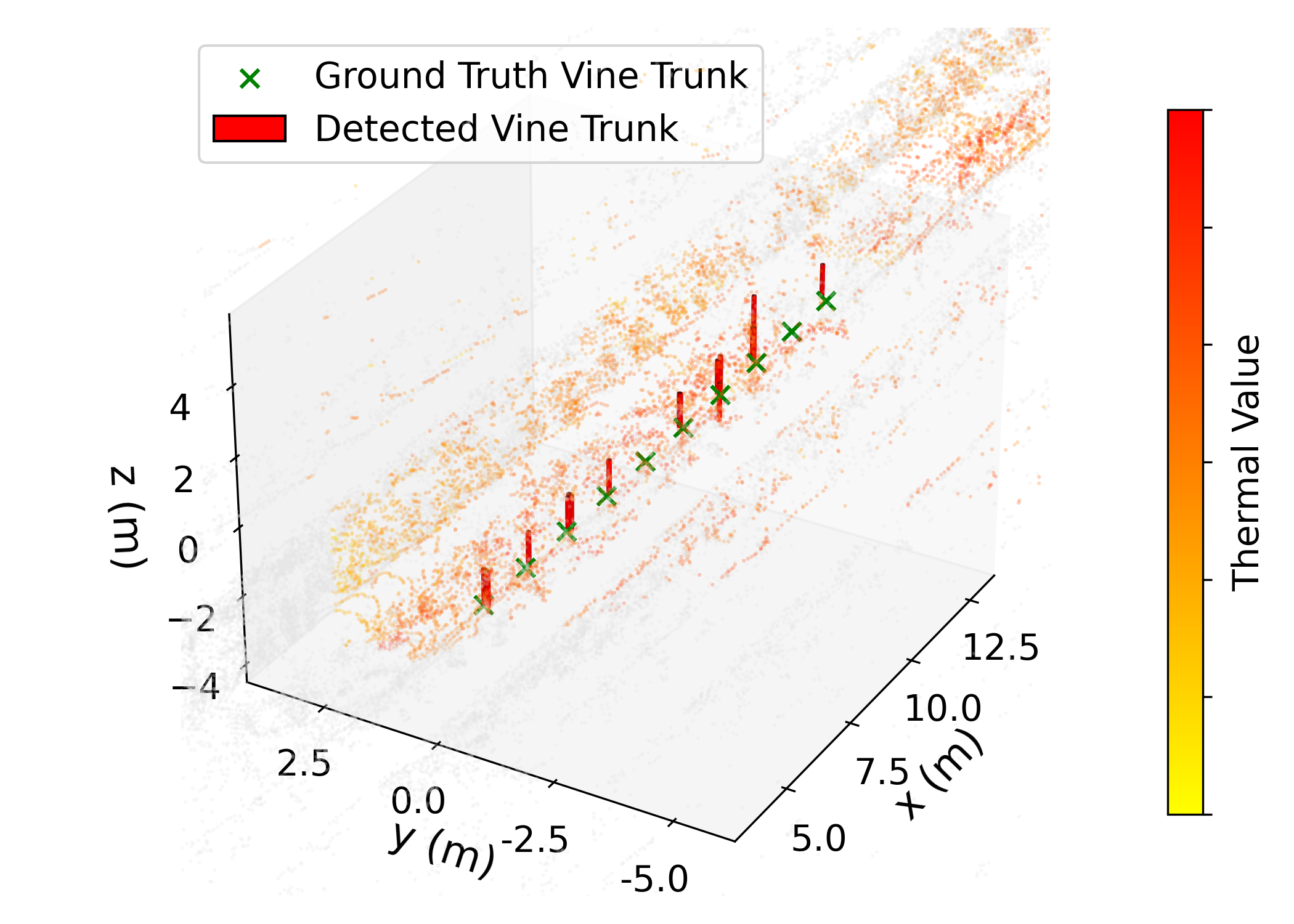}
\end{subfigure}%
\begin{subfigure}{.4\textwidth}
  \centering
  \includegraphics[width=1\linewidth]{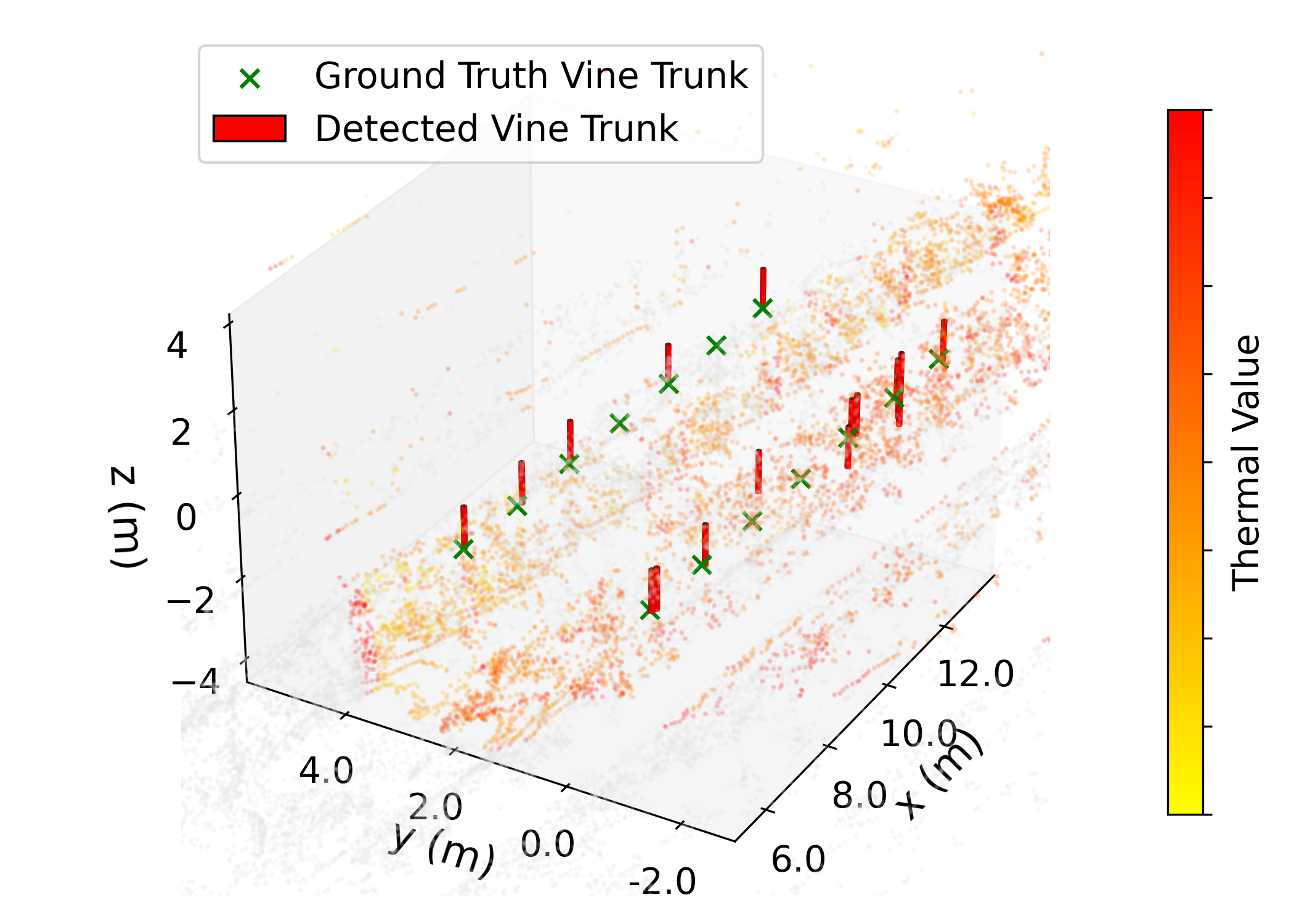}
\end{subfigure}%
\\
\begin{subfigure}{.42\textwidth}
  \centering
  \includegraphics[width=1\linewidth]{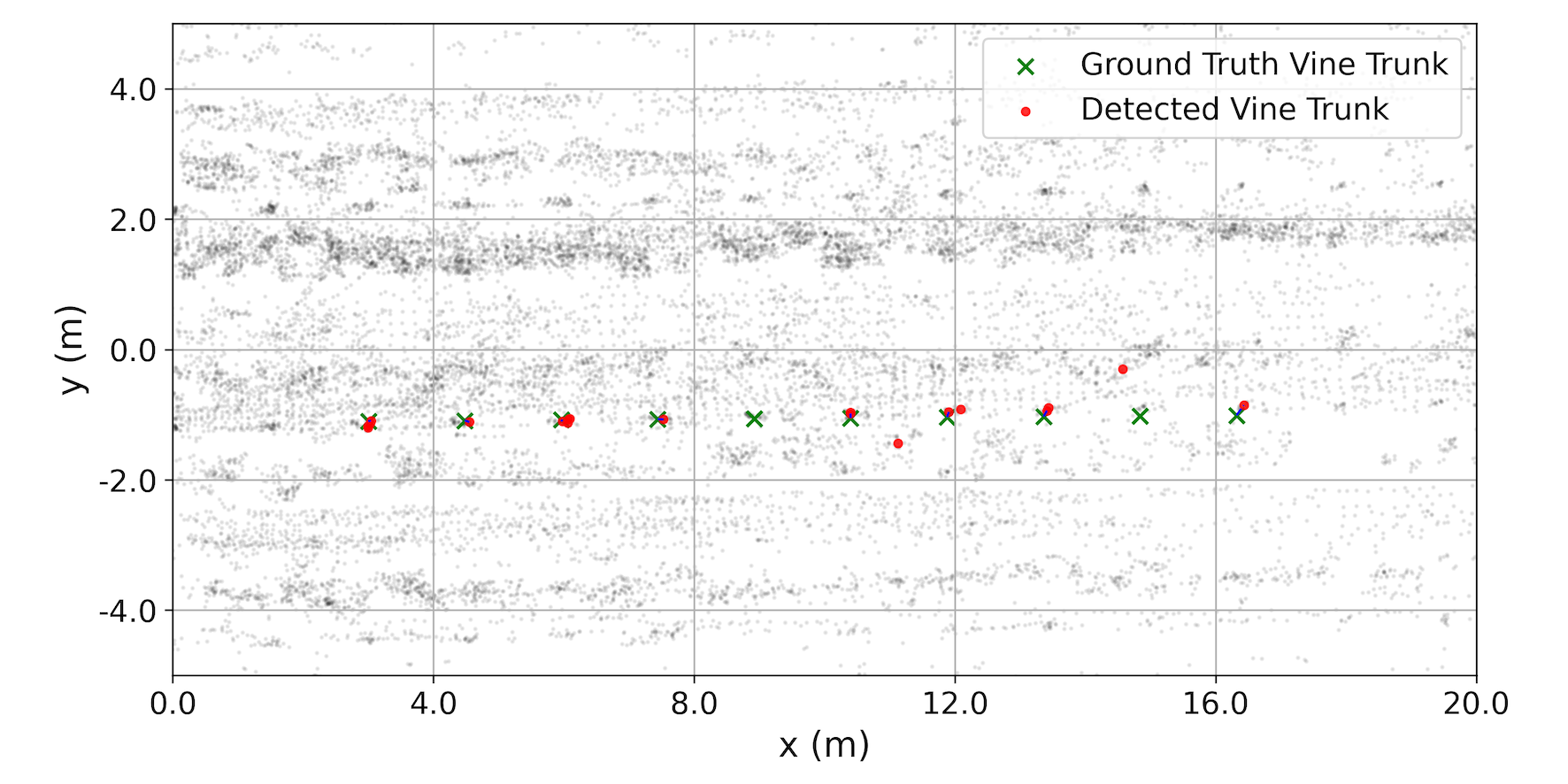}
\end{subfigure}
\begin{subfigure}{.42\textwidth}
  \centering
  \includegraphics[width=1\linewidth]
  {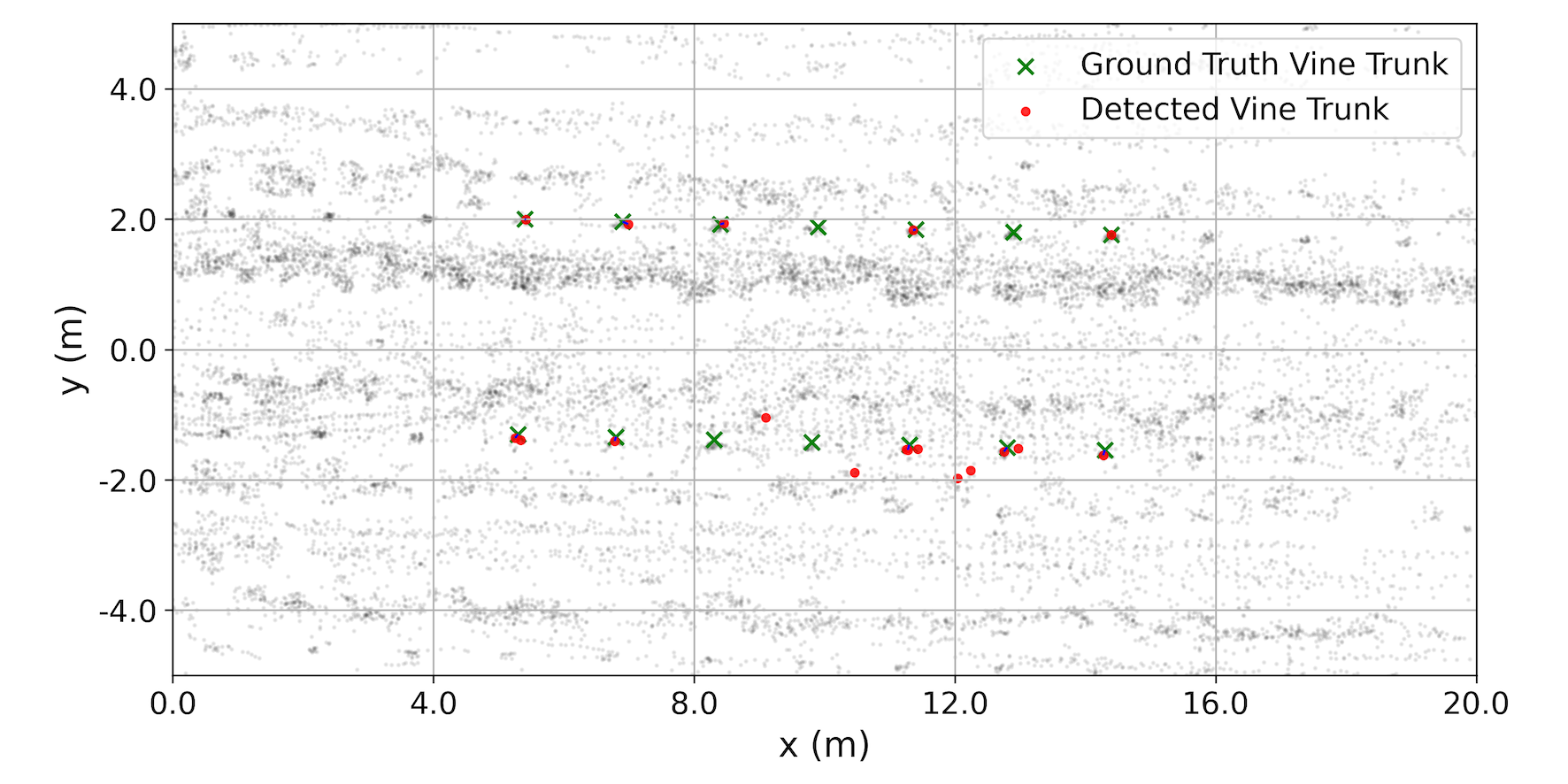}
\end{subfigure}
\caption{\textbf{Field assessment of vine detection and localization.} Our system accurately detects and localizes tree trunks in a single pass across both single-row (10 trees, left) and dual-row ($7\times7$ trees, right) configurations. By incorporating multi-modal information into the LOAM registration process, the system generates feature-rich sparse maps$-$such as the illustrated thermal visualizations$-$simultaneously with real-time vine trunk detection.}
\label{fig:detection_localisation_field_assesments}
\end{figure*}

Initially, as shown in Table~\ref{tab:field_evaluation}, our system was able to correctly detect vine trunks, identifying more than 80\% of the trees with a single pass from the robot. 
Specifically, in both the first and second sets of five trees, our detector demonstrated consistency scoring $0.32~m$ and $0.24~m$ of Euclidean ($L^2$) distance error with respect to the ground truth, respectively. 
Figure~\ref{fig:detection_localisation_field_assesments} illustrates tree trunk detections that are notably close to the ground truth tree positions. 
False-positive detections occurred when the trunk points fell in front of or behind the actual trunk points because of the deep vegetation around the observed vine, thus forming small clusters of false-positive points. 
The high recall rates across both five tree groups and the total 10 tree assessments demonstrate our system's reliability, exceeding 80\% in recall and minimizing the occurrence of false-negative trunk masks/detections. 
Notably, successful vine trunk detections were considered if they fell within a 3D Euclidean distance threshold of $0.15~m$ from the ground truth trunk positions.

\subsection{Dual-Row Vine Detection}

Following the evaluation setup of the single-vine row, we accessed our system on a $15\times5~m^2$ vineyard, where we enabled vine detections from the sides of the robot. In this way, the vine detector considers all appearing trees in the sensors' field of view during traverse. Herein, the detection task is more susceptible owing to the varying shadowing effects arising from the raycasted sunlight within vine rows  (seen also in Figures~\ref{fig:detectors_comparison_unseen}) during the post-noon survey hours.

As demonstrated by the results in Table~\ref{tab:field_evaluation}, our detector effectively acquires trunk detections by obtaining 10 out of 14 trees with a single field pass. Figure~\ref{fig:detection_localisation_field_assesments} also illustrates tree trunk detection in the local LOAM frame, along with the ground truth positions. 
Overall, our system achieves less than $0.37~m$ of $L^2$ error for each inspected tree by having less than $0.07~m$ of detection error measured in the Root Mean Squared Error (RMSE) in the candidate detections of $0.50~m$ radius surrounding the target trees. Through a single-pass field traversal and multi-modal detection of each trunk, our system demonstrates a consistently high recall rate$-$obtaining more than 70\% trees$-$even in the more variable sunlight assessment. 

\section{Conclusion}

This paper presents an onboard annotation-to-detection framework for identifying multi-modal objects of interest in the field, herein vine trunks, in scenarios where labeled data are limited. Our annotation pipeline leverages reliable and refined detectors to generate object pseudo-labels, which associate spatially across different modalities, to enrich a multi-modal training dataset. Simultaneously, the detection pipeline employs a multi-stage training procedure to iteratively refine the detector across different modalities, requiring minimal human intervention only in the early stage. Our system demonstrated robust vine trunk detection capabilities even when starting with a small curated dataset, obtaining more than 70\% of the trees in different vineyard settings with a single pass from the robot. 
Integration with LOAM and a tree trunk localization framework yielded a distance localization error below $0.37~m$, enabling the generation of multi-modal sparse maps.
The future implications of this work include scalable, multi-agent object detection that fuses cross-modal and cross-agent information to maximize detection accuracy and alleviate the need for human supervision during labor-intensive labeling tasks in the field.

\bibliographystyle{IEEEtran} 
\bibliography{refs}

\end{document}